\pdfoutput=1

\documentclass[11pt]{article}

\usepackage{acl}

\usepackage{url}
\usepackage{times}
\usepackage{latexsym}
\usepackage{hvfloat}
\usepackage{natbib}
\usepackage{makecell}
\usepackage{changes}
\usepackage{lipsum}
\usepackage{tikz}
\usetikzlibrary{shapes, arrows.meta, positioning}

\usepackage[T1]{fontenc}

\usepackage[utf8]{inputenc}

\usepackage{microtype}

\usepackage{inconsolata}

\usepackage{amsmath}
\usepackage{graphicx} 
\usepackage{amssymb}
\usepackage[edges]{forest}
\usepackage{booktabs}
\usepackage{multirow} 
\usepackage{todonotes}
\usepackage{changes}

\usepackage{graphicx}
\usepackage{tikz}
\usepackage{forest}

\usetikzlibrary{trees,positioning,shapes,shadows,arrows.meta}

\definecolor{lightcoral}{rgb}{0.94, 0.5, 0.5}
\definecolor{lightgreen}{rgb}{0.56, 0.93, 0.56}
\definecolor{harvestgold}{rgb}{0.98, 0.85, 0.40}
\definecolor{brightlavender}{rgb}{0.75, 0.58, 0.89}
\definecolor{capri}{rgb}{0.0, 0.75, 1.0}
\definecolor{carminepink}{rgb}{0.92, 0.3, 0.26}
\definecolor{celadon}{rgb}{0.67, 0.88, 0.69}
\definecolor{darkpastelgreen}{rgb}{0.01, 0.75, 0.24}

\definecolor{hidden-draw}{RGB}{205, 44, 36}
\definecolor{hidden-blue}{RGB}{194,232,247}
\definecolor{hidden-orange}{RGB}{243,202,120}
\definecolor{hidden-yellow}{RGB}{242,244,193}
\definecolor{tree-level-1}{RGB}{245,20,85}
\definecolor{tree-level-2}{RGB}{246,86,118}
\definecolor{tree-level-3}{RGB}{248,177,193}
\definecolor{tree-leaf}{RGB}{176,230,198}

\definecolor{Self}{RGB}{255,0,128}
\definecolor{Ensemble}{RGB}{0,127,255}
\definecolor{Iterative}{RGB}{153,51,255}

\definecolor{exemplar1}{RGB}{136,98,148}
\definecolor{exemplar2}{RGB}{148,210,242}
\definecolor{knowledge1}{RGB}{249,219,152}
\definecolor{knowledge2}{RGB}{255,245,220}

\newcommand{\commentisunseen}{0}

\newcommand{\commentp}[1]{
}

\newcommand{\fc}[1]{
\ifthenelse{\equal{\commentisunseen}{0}}{
{\color{blue}#1}}
{#1}
}
\newcommand{\FC}[1]{
\ifthenelse{\equal{\commentisunseen}{0}}{
{\color{blue}FC: #1}}
{}
}

\DeclareMathOperator*{\argmax}{arg\,max}

\title{A Survey of Confidence Estimation and Calibration\\ in Large Language Models}

\author{
${\textbf{Jiahui Geng}}^1$,
${\textbf{Fengyu Cai}}^2$,
${\textbf{Yuxia Wang}}^1$, \\
${\textbf{Heinz Koeppl}}^2$,
${\textbf{Preslav Nakov}}^1$,
${\textbf{Iryna Gurevych}}^{1}$  \\
$^1$ Mohamed bin Zayed University of Artificial Intelligence \\
$^2$ Technical University of Darmstadt\\
\{jiahui.geng, yuxia.wang,preslav.nakov,iryna.gurevych\}@mbzuai.ac.ae, \\
\{fengyu.cai,heinz.koeppl\}@tu-darmstadt.de
}

\begin{document}
\maketitle
\begin{abstract}
Large language models (LLMs) have demonstrated remarkable capabilities across a wide range of tasks in various domains. Despite their impressive performance, they can be unreliable due to factual errors in their generations. Assessing their confidence and calibrating them across different tasks can help mitigate risks and enable LLMs to produce better generations. There has been a lot of recent research aiming to address this, but there has been no comprehensive overview to organize it and outline the main lessons learned. The present survey aims to bridge this gap. In particular, we outline the challenges and we summarize recent technical advancements for LLM confidence estimation and calibration. We further discuss their applications and suggest promising directions for future work.  
\end{abstract}

\section{Introduction}
Large language models (LLMs) have demonstrated a wide range of capabilities, such as world knowledge storage, sophisticated language-based reasoning, and in-context learning~\cite{petroni2019language, wei2022chain,brown2020language}. However, LLMs do not consistently achieve good performance~\cite{factsurveywang2023,zhang2023siren}. Their generation still includes biases~\cite{zhao2021calibrate,wang2023reducing} and hallucinations that do not align with reality~\cite{zhang2023siren}. Evaluating the trustworthiness of responses from these models remains challenging~\cite{liu2023trustworthy}.


Confidence (or uncertainty) estimation is crucial for tasks like out-of-distribution detection and selective prediction~\cite{DBLP:conf/nips/KendallG17,lu2022learning}, and it has been extensively studied and applied in various contexts ~\cite{lee2018simple,devries2018learning}. A related concept is that of model calibration, which focuses on aligning predictive probabilities (estimated confidence) to actual accuracy~\cite{Guo2017OnCO}. 

However, applying these methods directly to LLMs presents several challenges. The output space of these models is significantly larger than that of discriminative models. The number of possible outcomes grows exponentially with the generation length, making it impossible to access all potential responses. Additionally, different expressions may convey the same meaning, suggesting that confidence estimation should consider semantics.
{Lastly,} LLMs show unique properties, such as expressing confidence in words~\cite{lin2022teaching,xiong2023can} and the ability to perform zero-shot or few-shot learning~\cite{brown2020language}. Nonetheless, their responses can be sensitive to the prompts, e.g., the examples provided and their order, which can cause a lot of instability in the results. Given this, confidence estimation and calibration for LLMs is growing as an emerging area of interest~\cite{jiang2021can,lin2022teaching,lin2023generating,shrivastava2023llamas}. 

While existing surveys mainly focused on issues such as hallucination and factuality in LLMs~\cite{zhang2023siren,wang2023survey}, there are no comprehensive surveys systematically discussing the technical advancements in LLMs, and here we aim to bridge this gap. We explore the unique challenges posed by LLMs and examine the latest studies addressing these issues. We first discuss key concepts such as confidence, uncertainty, and calibration in the context of neural models, as detailed in Section~\ref{sec:preliminary}. Then, we pursue two different directions: one addressing confidence estimation and calibration techniques for generation tasks in Section~\ref{sec:gm4ge}, and the other for classification tasks in Section~\ref{sec:gm4cl}. We conclude by exploring their practical applications (Section~\ref{sec:applications}) and looking at potential future research directions (Section~\ref{sec:future_directions}).  {Figure~\ref{categorization_of_survey} provides a comprehensive representation of the survey's structure. By conducting a detailed examination of existing research, our goal is to illuminate this vital facet of LLMs, contributing to the development of more reliable applications.}

\tikzstyle{my-box}=[
    rectangle,
    draw=hidden-draw,
    rounded corners,
    text opacity=1,
    minimum height=1.5em,
    minimum width=5em,
    inner sep=2pt,
    align=center,
    fill opacity=.5,
]
\tikzstyle{cause_leaf}=[my-box, minimum height=1.5em,
    fill=harvestgold!20, text=black, align=left,font=\scriptsize,
    inner xsep=2pt,
    inner ysep=4pt,
]
\tikzstyle{detect_leaf}=[my-box, minimum height=1.5em,
    fill=cyan!20, text=black, align=left,font=\scriptsize,
    inner xsep=2pt,
    inner ysep=4pt,
]
\tikzstyle{mitigate_leaf}=[my-box, minimum height=1.5em,
    fill=lightgreen!20, text=black, align=left,font=\scriptsize,
    inner xsep=2pt,
    inner ysep=4pt,
]

\begin{figure*}[tp]
    \centering
    \resizebox{\textwidth}{!}{
        \begin{forest}
            forked edges,
            for tree={
                grow=east,
                reversed=true,
                anchor=base west,
                parent anchor=east,
                child anchor=west,
                base=left,
                font=\small,
                rectangle,
                draw=hidden-draw,
                rounded corners,
                align=left,
                minimum width=4em,
                edge+={darkgray, line width=1pt},
                s sep=3pt,
                inner xsep=2pt,
                inner ysep=3pt,
                ver/.style={rotate=90, child anchor=north, parent anchor=south, anchor=center},
            },
            [
                Confidence Estimation and Calibration in LLMs, ver, color=carminepink!100, fill=carminepink!15,
                text=black
                [
                    Metrics, color=harvestgold!100, fill=harvestgold!100, text width=4.0em, text=black
                    [
                        Classification, color=harvestgold!100, fill=harvestgold!60,  text width=5em, text=black
                        [
                            \citet{Guo2017OnCO,nixon2019measuring,classwiseece,auroc}, cause_leaf, text width=23.5em
                        ]
                    ]
                    [
                        Generation, color=harvestgold!100, fill=harvestgold!60, text width=5em, text=black
                        [
                            \citet{kumar2019calibration,lin2023generating,zhu2023calibration,huang2024calibrating}, cause_leaf, text width=23.5em
                        ]   
                    ]   
                ]
                [
                    Methods, color=cyan!100, fill=cyan!100, text width=4.0em, text=black
                    [
                        Generation, color=cyan!100, fill=cyan!80, text width=5em, text=black
                        [
                            Estimation, color=cyan!100, fill=cyan!60, text width=3.5em, text=black
                            [
                                Logit-based methods, color=cyan!100, fill=cyan!30, text width=7.0em, text=black
                                    [
                                         {\citet{duan2023shifting,kuhn2023semantic}}
                                        , detect_leaf, text width=20em
                                    ]
                            ]
                            [
                                Internal state-based \\ methods, color=cyan!100, fill=cyan!30, text width=7.0em, text=black
                                    [
                                        {\citet{ren2022out,kadavath2022language,burns2023discovering}\\ \citet{li2023inference,azaria2023internal}}
                                        , detect_leaf, text width=20em
                                    ]
                            ]
                            [
                                Linguistic confidence, color=cyan!100, fill=cyan!30, text width=7.0em, text=black
                                    [
                                        {\citet{mielke2022reducing,xiong2023can}}
                                            , detect_leaf, text width=20em
                                    ]
                            ]
                            [
                                Consistency-based \\ estimation, color=cyan!100, fill=cyan!30, text width=7.0em, text=black
                                    [
                                        {\citet{manakul2023selfcheckgpt,lin2023generating}}
                                        , detect_leaf, text width=20em
                                    ]
                            ]
                            [
                                Surrogate models, color=cyan!100, fill=cyan!30, text width=7.0em, text=black
                                [
                                    {\citet{shrivastava2023llamas,touvron2023llama}}
                                    , detect_leaf, text width=20em
                                ]
                            ]
                        ]
                        [
                            Calibration, color=cyan!100, fill=cyan!60, text width=3.5em, text=black
                            [
                                Improving generation, color=cyan!100, fill=cyan!30, text width=7.0em, text=black
                                [
                                    {\citet{kumar2019calibration,wang2020inference,lu2022learning}\\ \citet{xiao2021hallucination,van2022mutual,zablotskaia2023uncertainty}\\ \citet{zhao2022calibrating,zhao2023slic}}
                                    , detect_leaf, text width=20em
                                ]
                            ]
                            [
                                Improving linguistic \\ confidence, color=cyan!100, fill=cyan!30, text width=7.0em, text=black
                                [
                                    {\citet{mielke2022reducing,lin2022teaching,zhou2023navigating}}
                                    , detect_leaf, text width=20em
                                ]
                            ]
                        ]
                    ]
                    [
                        Classification, color=cyan!100, fill=cyan!80, text width=5em, text=black
                        [
                            Estimation, color=cyan!100, fill=cyan!60, text width=3.5em, text=black
                            [
                                Logit-based method, color=cyan!100, fill=cyan!30, text width=7.0em, text=black
                                [
                                    {\citet{mielke2022reducing,lin2022teaching,zhou2023navigating}}
                                    , detect_leaf, text width=20em
                                ]
                            ]
                        ]
                        [
                            Calibration, color=cyan!100, fill=cyan!60, text width=3.5em, text=black
                            [
                                Bias mitigation, color=cyan!100, fill=cyan!30, text width=7.0em, text=black
                                [
                                    {\citet{zhao2021calibrate,fei2023mitigating,nie2022improving,han2022prototypical}}
                                    , detect_leaf, text width=20em
                                ]
                            ]
                        ]
                    ]
                ]
                [
                    Application, color=lightgreen!100, fill=lightgreen!100, text width=4.0em, text=black
                       [
                            Hallucination Detection
                            \\ and Mitigation, color=lightgreen!100, fill=lightgreen!60, text width=8em, text=black
                            [
                                {\citet{manakul-etal-2023-cued,Zhang2023SAC3RH}\\ \citet{varshney2023stitch}}
                                        , mitigate_leaf, text width=14em
                            ]
                       ]
                       [
                            Ambiguity detection \\ and selective generation, color=lightgreen!100, fill=lightgreen!60, text width=8em, text=black
                            [  {\citet{kamath2020selective,zablotskaia2023uncertainty}\\ \citet{cole2023selectively,hou2023decomposing}}
                                        , mitigate_leaf, text width=14em
                            ]
                       ]
                       [
                            Uncertainty-guided data  \\ exploitation , color=lightgreen!100, fill=lightgreen!60, text width=8em, text=black
                            [
                                {\citet{yu2022cold,su2022selective,jiang2023active}}
                                        , mitigate_leaf, text width=14em
                            ]
                        ]
                ]
            ]   
        \end{forest}
    }
\caption{The taxonomy of confidence estimation and calibration in LLMs.}
\label{categorization_of_survey}
\end{figure*}
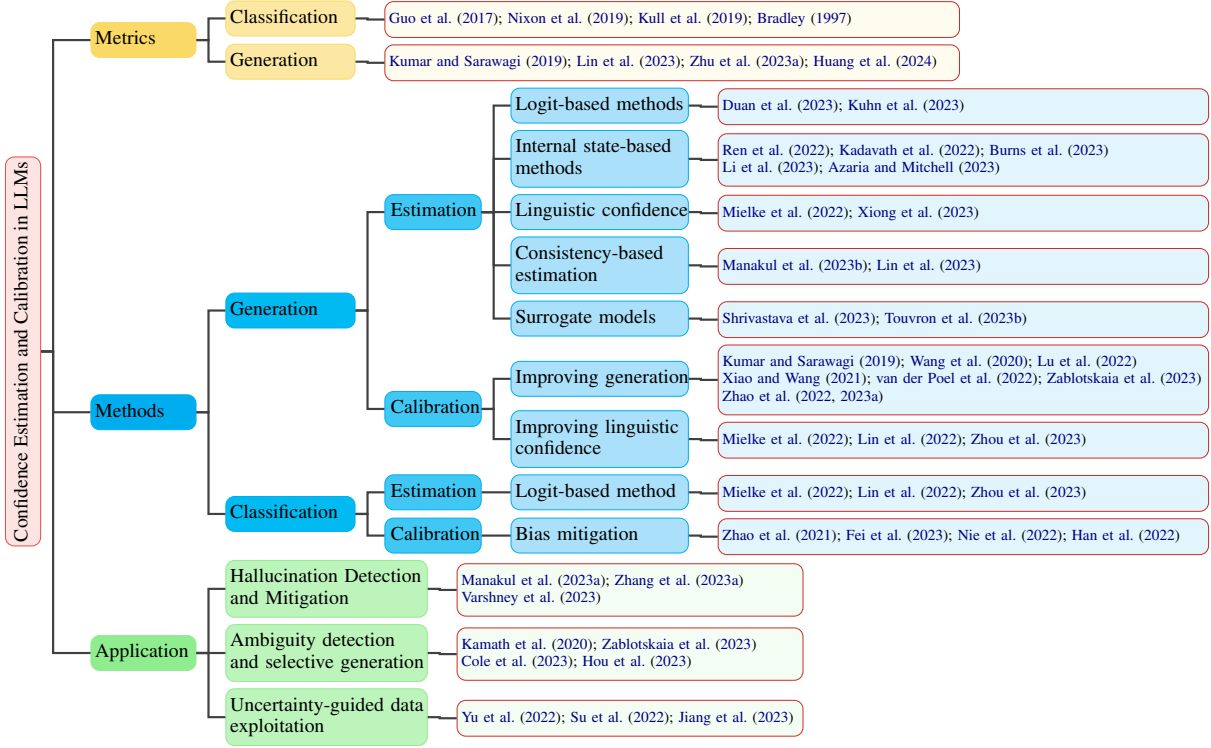

\section{Preliminaries and Background}
\label{sec:preliminary}
\subsection{Basic Concepts}

In machine learning, confidence and uncertainty are two facets of a single principle: higher confidence corresponds to lower uncertainty~\cite{xiao2022uncertainty,chen2023quantifying}. Research on quantifying model confidence has led to the development of two key concepts: \textit{relative confidence score} and \textit{absolute confidence score}, offering different methods to assess and to interpret confidence levels~\cite{kamath2020selective,vazhentsev2023hybrid}. Given input $x$, ground truth $y$, and prediction $\hat{y}$, the model's predictive confidence is denoted as $\texttt{conf}(x, \hat{y})$. Relative confidence scores emphasize the ability to rank samples, distinguishing correct predictions from incorrect ones. Ideally, for every pair of $(x_i, y_i)$ and $(x_j, y_j)$ and their corresponding predictions $\hat{y}_i$ and $\hat{y}_j$, we have
    \begin{equation}
    \label{eq:odranking}
        \begin{split}
                    \texttt{conf}(\mathbf{x}_i, \hat{y}_i) & \leq \texttt{conf}(\mathbf{x}_j, \hat{y}_j) \\
        \Longleftrightarrow
        P(\hat{y}_i = y_i| \mathbf{x}_i) & \leq P(\hat{y}_j = y_j| \mathbf{x}_j)
        \end{split}
    \end{equation}

An absolute confidence score indicates that a model's score reflects its true accuracy in real-world scenarios. For example, if a model predicts an event with a $70\%$ probability, that event should actually happen about $70\%$ of the time under similar circumstances. The equation for this relationship is as follows:
    \begin{equation}
        P(\hat{y} = y \mid \texttt{conf}(x, \hat{y}) = q) = q
    \label{eq:calibration-definition}
    \end{equation}
When the model's predicted confidence scores consistently align with this principle, the model is considered to be well-calibrated.

 \citet{DBLP:conf/nips/KendallG17} proposed categorizing uncertainty in machine learning into \emph{aleatoric} and \emph{epistemic} uncertainty. Aleatoric or data uncertainty emerges from the inherent randomness or variability of a system or a process. It is an intrinsic feature of the system and is typically irreducible. Epistemic uncertainty, in contrast, is known as model uncertainty or systematic uncertainty. It arises from the lack of knowledge or information about the system being modeled and is reducible, as it can diminish with the acquisition of more data and improved modeling techniques~\cite{DBLP:conf/icml/GalG16,DBLP:conf/nips/Lakshminarayanan17}.

\subsection{Metrics and Methods}
\paragraph{Metrics} Due to the continuous nature of confidence scores, it is impossible to accurately calculate the probability as in Eq.~\ref{eq:calibration-definition}. Expected calibration error (ECE;~\citealt{Guo2017OnCO}) approximates it by clustering instances with similar confidence. The predicted probabilities are first segmented into various bins. ECE is then calculated by taking the weighted average of the discrepancies between the mean predicted probability and the actual accuracy across all bins $B_m, m=1\cdots, M$: 
\begin{equation}
    ECE = \sum_{m=1}^M  \frac{\lvert B_m\rvert}{N} \lvert acc(B_m) - conf(B_m) \rvert 
\end{equation}
One drawback of the ECE metric is its sensitivity to various factors such as bucket width and the variance of samples within these buckets. To overcome these issues, more sophisticated schemes have been developed, including static calibration error (SCE), adaptive calibration error (ACE; \citealt{nixon2019measuring}), and classwise ECE~\cite{classwiseece}. ECE can also be visualized as a reliability diagram, which plots predicted probabilities against observed frequencies, with points or lines above the diagonal indicating overconfidence. Additionally, metrics such as F1 score, {area under receiver operating characteristic curve} (AUROC; \citealt{auroc}) and {area under accuracy-rejection curve} (AUARC; \citealt{lin2023generating}), can indicate whether the confidence score can appropriately differentiate between correct and incorrect answers.

{However, it's necessary to adapt metrics to effectively process sequence of tokens with semantics. A common approach is to evaluate whether the next token probability is well-calibrated. Assuming that $\mathbf{y}_i = y_{i1}, \cdots, y_{iT}$ denotes the sequence of generated tokens (target sentence) and that $\mathbf{x}_i = x_{i1}, \cdots, x_{iS}$ denotes the sequence of input tokens (source sentence), the probability of generating the target sequence can be represented as: $\prod_{t=1}^{T} P(y_{it}|\mathbf{x}_i, \mathbf{y}_{i,<t})$. For simplicity, we use $P_{it}(y_{it})$ to represent $P(y_{it}|\mathbf{y}_{i, <t}, \mathbf{x}_i)$ and $C_{it}(y) = \delta(y_{it} = y)$ to denote if $y$ matches the correct label $y_{it}$}. The ECE can be mathematically expressed as:
\begin{equation}
     \frac{1}{L} \sum_{m=1}^{M} \lvert \sum_{i,t:P_{it}(\hat{y}_{it}) \in B_m} C_{it}(\hat{y}_{it}) - P_{it}(\hat{y}_{it})\rvert
\end{equation}
where $L = \sum_{i=1}^N \lvert \mathbf{y}_i \rvert $ is the total number of generated tokens.
{\citet{kumar2019calibration} claimed that such metric focuses solely on the highest score label, neglecting the entire probability distribution, and thereby introduced \textit{weighted ECE} for refined calibration distinction. Another approach analyzes the overall correctness and confidence of answers directly, especially in tasks like classification and question answering~\cite{lin2022teaching,kadavath2022language}. \citet{huang2024calibrating} treated correctness as distributions instead of binary values, assessing calibration through the distance between correctness and confidence. }


\paragraph{Methods in discriminative models} Common methods for confidence estimation include logit-based methods~\cite{pearce2021understanding,pereyra2017regularizing}, ensemble-based and Bayesian methods~\cite{DBLP:conf/nips/Lakshminarayanan17,DBLP:conf/icml/GalG16}, density-based methods~\cite{lee2018simple}, and confidence-learning methods~\cite{devries2018learning}. Model calibration~\cite{Guo2017OnCO} can either occur during the model's training phase, for example, by improving loss functions~\cite{szegedy2016rethinking} or be applied after the model has been trained, such as temperature scaling (TS;~\citealt{Guo2017OnCO}) and feature-based calibrators (FBC;~\citealt{jiang2021can}).  Table~\ref{tab:genlms1} represents significant research in the discriminative LMs, with a list of models, tasks, and calibration methods. Due to space limitations, please refer to the 
 Appendix~\ref{sec:appendix} for detailed principles and comparisons.

\begin{table*}[htp]
\centering
\resizebox{\textwidth}{!}{
\begin{tabular}{p{4cm}p{5cm}p{13cm}}
\toprule
\textbf{Study} & \textbf{Model} &  \textbf{Proposed Methods} \\ 
\midrule
\citet{duan2023shifting} & OPT~\cite{zhang2022opt}  & SAR (Shifting Attention to Relevance): consider semantic relevance when evaluating token and sentence-level uncertainty   \\ 
\citet{manakul2023selfcheckgpt} & GPT-3~\cite{NEURIPS2020_1457c0d6} & Semantic uncertainty: evaluate the consistency of responses by various methods \\
\citet{kuhn2023semantic} & OPT~\cite{zhang2022opt} & 
Cluster answers according to semantics and then computes the sum of probabilities within each cluster to represent confidence  \\
\citet{kadavath2022language} & Anthropic LLM~\cite{bai2022training} & P(True): the probability a model assigns to its answer as True, P(IK): probability a model assigns to "I know" by leveraging a binary classifier \\
\citet{xiong2023can} & \makecell[l]{GPT3/3.5/4~\cite{NEURIPS2020_1457c0d6}, \\Vicuna~\cite{vicuna2023}} & Hybrid methods combining linguistic confidence and consistency-based confidence\\
\citet{lin2023generating} & GPT-3.5 & Estimate confidence by evaluating the lexical and semantic similarity among responses\\
\citet{shrivastava2023llamas} & GPT-3.5/4, Claude & Hybrid methods combing confidence from surrogate models and linguistic confidence of target models \\
\bottomrule
\end{tabular}}
\caption{\textbf{Recent studies of LLM confidence estimation}. 
These studies evaluate confidence estimation in question-answering tasks, utilizing  metrics such as ECE, AUROC, etc.}
\label{tab:estimation}
\end{table*}



\begin{table*}[ht]
\centering
\resizebox{\textwidth}{!}{
\begin{tabular}{p{4cm}p{6cm}p{6cm}p{6cm}}
\toprule
\textbf{Study} & \textbf{{Model}} & \textbf{Task} & \textbf{Calibration Methods} \\ 
\midrule
\citet{kumar2019calibration}    & \makecell[l]{LSTM~\cite{DBLP:journals/corr/BahdanauCB14}, \\ Transformer~\cite{NIPS2017_3f5ee243}}    &Machine Translation   &  TS with Learnable Parameters                        \\ 
\citet{lu2022learning}     &   Transformer~\cite{NIPS2017_3f5ee243} &  Machine Translation   &  Confidence-Based LS                             \\ 
\citet{wang2020inference}     &  Transformer~\cite{NIPS2017_3f5ee243}   & Machine Translation   &      LS, Dropout    \\ 
\citet{xiao2021hallucination}      &   \makecell[l]{LSTM~\cite{DBLP:journals/corr/BahdanauCB14}, \\ Transformer~\cite{NIPS2017_3f5ee243}}  & \makecell[l]{Data2Text Generation, \\Image Captioning }   &      Uncertainty-Aware Decoding                     \\ 
\citet{van2022mutual}      &   BART~\cite{lewis2019bart}  &  Text Summarization   &    CPMI-Based Decoding                     \\ 
\citet{zablotskaia2023uncertainty}      &   T5~\cite{raffel2020exploring}  & Text Summarization   &   \makecell[l]{MC-Dropout, BE, SNGP,\\ DeepEnsemble }                     \\ 
\citet{zhao2022calibrating}      &   PEGASUS~\cite{zhang2020pegasus}  & \makecell[l]{ Text Summarization, \\Question Answering}     &    SLiC 
\\ 
\citet{zhao2023slic}      &   T5~\cite{raffel2020exploring}  & Text Summarization  &    SLiC-HF
\\ 
\citet{mielke2022reducing}      &   BlenderBot~\cite{roller-etal-2021-recipes} &  Dialogue Generation   &    Linguistic Calibration
\\ 
\citet{lin2022teaching}      &   GPT-3~\cite{NEURIPS2020_1457c0d6} & Math Question Answering  &   Fine-Tuning
\\ 
\midrule
\citet{zhao2021calibrate} & GPT-3~\cite{NEURIPS2020_1457c0d6} & \makecell[l]{Text Classification, Fact Retrieval \\ Information Extraction} & Contextual Calibration 
\\ 
\citet{fei2023mitigating} & \makecell[l]{PALM-2~\cite{DBLP:journals/corr/abs-2305-10403},\\ CLIP~\cite{DBLP:conf/icml/RadfordKHRGASAM21}}  & Text Classification  & Domain-Context Calibration 
\\ 
\citet{han2022prototypical} & GPT-2~\cite{radford2019language} &  Text Classification & Prototypical Calibration
\\ 
\citet{kumar2022answer} & GPT-2~\cite{radford2019language} & Multiple Choice Question Answering & Answer-Level Calibration 
\\ 
\citet{holtzman2021surface} & \makecell[l]{GPT-2\cite{radford2019language}, \\GPT-3~\cite{NEURIPS2020_1457c0d6}} & Multiple Choice Question Answering & PMIDC 
\\ 
\citet{DBLP:journals/corr/abs-2309-03882} & \makecell[l]{LLaMA~\cite{DBLP:journals/corr/abs-2302-13971},\\ Vicuna~\cite{vicuna2023},\\ Falcon~\cite{DBLP:journals/corr/abs-2306-01116}, GPT-3.5} & Multiple Choice Question Answering & PriDE 
\\ 
\bottomrule

\end{tabular}}
\caption{\textbf{Studies of LLM calibration}. The first half is about generation tasks, and the second half is about classification tasks. \textbf{Calibration methods:}  LS: label smoothing, TS: temperature scaling, BE: Bayesian ensemble, SNGP: spectral-normalized Gaussian process, MCDropout: Monte Carlo dropout, SLiC: sequence likelihood calibration, HF: human feedback, FBC: feature-based calibrator, CPMI: conditional pointwise mutual information, PMIDC: domain conditional pointwise mutual information, PriDE: debiasing with prior estimation.}
\label{tab:genlms}
\end{table*}

\section{LLMs for Generation Tasks}
\label{sec:gm4ge}

\subsection{Confidence Estimation}
In this section, we generally divide the methods into white-box and black-box methods. We first provide a detailed overview of these methods and then summarize their strengths, weaknesses, and connections.

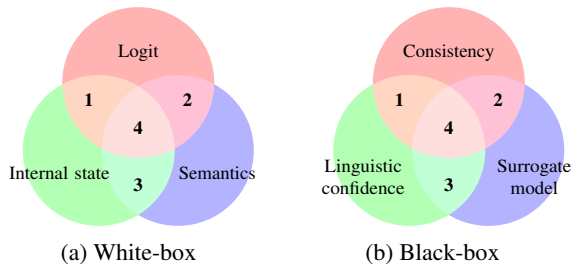
\begin{figure}[!htp]
\scriptsize 
\begin{subfigure}[b]{0.25\textwidth}
    \begin{center}
        \begin{tikzpicture}[scale=0.5]
            \begin{scope}[blend group = soft light]
            \fill[red!30!white]   ( 90:1.2) circle (2);
            \fill[green!30!white] (210:1.2) circle (2);
            \fill[blue!30!white]  (330:1.2) circle (2);
            \end{scope}
            \node at ( 90:2)    {Logit };
            \node at ( 210:2.4) [text width=1.5cm, align=center] {Internal  state};
            \node at ( 330:2.4)   {Semantics};
            \node at ( 150:1.5)   [align=center]{\textbf{1}};
            \node at ( 30:1.5)    [align=center]{\textbf{2}};
            \node at ( 270:1.5)   [align=center] {\textbf{3}};
            \node [align=center] {\textbf{4}};
        \end{tikzpicture}
        \subcaption{White-box}
        \label{fig:white}
    \end{center}
\end{subfigure}\begin{subfigure}[b]{0.25\textwidth}
    \begin{center}
        \begin{tikzpicture}[scale=0.5]
            \begin{scope}[blend group = soft light]
            \fill[red!30!white]   ( 90:1.2) circle (2);
            \fill[green!30!white] (210:1.2) circle (2);
            \fill[blue!30!white]  (330:1.2) circle (2);
            \end{scope}
            \node at ( 90:2)    {Consistency};
            \node at ( 210:2.6) [text width=2cm, align=center]  {Linguistic confidence};
            \node at ( 330:2.6) [text width=1.5cm, align=center]  {Surrogate model};
            \node at ( 150:1.5)   [align=center]{\textbf{1}};
            \node at ( 30:1.5)    [align=center]{\textbf{2}};
            \node at ( 270:1.5)   [align=center] {\textbf{3}};
            \node [align=center] {\textbf{4}};        
        \end{tikzpicture}
        \subcaption{Black-box}
    \label{fig:black}
    \end{center}
\end{subfigure}
\caption{Venn diagram: the taxonomy of information sources for white-box (\textbf{Left}) and black-box (\textbf{Right}) confidence estimation methods. These two families of methods can be categorized into the methods relying on logit, internal state, or semantics, and those relying on consistency, linguistic confidence, or surrogate model, respectively.
The intersections of these methods are located in Zone \textbf{1} - \textbf{4}. }
\label{fig:taxonomy}
\end{figure} 



 
\subsubsection{White-Box Methods}

 White-box methods operate on the premise that the state at every position of the LLMs is accessible during inference. 
\paragraph{Logit-based methods} The logit-based method evaluates sentence uncertainty using token-level probabilities or entropy~\cite{huang2023look}. To ensure an evaluation consistent across sentences of different lengths, the length-normalized likelihood probability is widely utilized \cite{DBLP:conf/wmt/MurrayC18}. Moreover, alternatives such as the minimum or average token probabilities and the average entropy are also widely used ~\cite{vazhentsev2023efficient}. Logit-based techniques readily adapt to scenarios involving multiple sampling~\cite{vazhentsev2023efficient} or ensemble models~\cite{DBLP:conf/iclr/MalininG21}. 

To incorporate semantics,~\citet{duan2023shifting} introduced the concept of \emph{token-level relevance}, which evaluates the relevance of the token by comparing semantic change before and after moving the token with a semantic similarity metric like Sentence Transformer~\cite{DBLP:conf/emnlp/ReimersG19}. Then, sentence uncertainty can be adjusted based on the token's relevance. ~\citet{duan2023shifting} further proposed ~\emph{sentence-level relevance} in multiple sampling settings, considering the similarity between the returned sentence and other sampled ones.
 \citet{kuhn2023semantic} proposed \textit{semantic uncertainty}, which first clusters semantically equivalent samples based on the bidirectional entailment between samples and then approximates semantic entropy by summing probabilities in each cluster.

\citet{kadavath2022language} discovered that LLMs can self-assess to differentiate between correct and incorrect answers. They suggested a method called \textit{P(True)}, where the LLM first generates responses and then evaluates them as "True" or "False". The probability the model assigns the confidence level to "True” determines the confidence level.

\paragraph{Internal state-based methods}
\citet{ren2022out} introduced a technique for out-of-distribution detection and selective generation. The method starts by computing embeddings for both inputs and outputs in the training data, fitting them to a Gaussian distribution. It then assesses the model's confidence in its generated data by calculating the relative Mahalanobis distance of the evaluated data pair from this Gaussian distribution.

Recent studies have posited the existence of a direction in activation space that effectively separates true and false inputs~\cite{kadavath2022language,burns2023discovering,li2023inference,azaria2023internal}. \citet{kadavath2022language} proposed training a classifier (the probe), named P(IK), on the activations of a network to predict whether an LLM knows the answer. They sampled multiple answers for each question at a consistent temperature, labeled the correctness of each answer, and then used the question-correctness pair as the training data. Similarly,~\citet{li2023inference} and \citet{azaria2023internal} employed linear probes to examine whether attention heads in various layers can differentiate between correct and incorrect answers. Their empirical findings indicated that certain middle layers and a few attention heads exhibit strong performance in this task, although the layer positions vary across models. ~\citet{burns2023discovering} introduced an unsupervised approach to map hidden states to probabilities. It entails responding to questions with "Yes" or "No," extracting and converting model activations into truth probabilities, and optimizing unsupervised loss for consistency. It ultimately gauges the model's confidence by estimating the likelihood of a "Yes" response. 



\paragraph{Summary} 

White-box methods, as illustrated in Figure~\ref{fig:white}, primarily utilize logits, internal states, and semantics as sources of information. Logit-based approaches, easy to implement during inference, face a limitation in that low logit probabilities may reflect various properties of language. Methods focusing on internal states~\cite{kadavath2022language,li2023inference,azaria2023internal} provide insights into the model's linguistic understanding, though they typically require supervised training on specially annotated data. \citet{levinstein2024still} highlighted the limitations of the probing method in generalizing to unseen examples with negations. Semantics are often used to complement other methods, providing them with interpretability~\cite{kuhn2023semantic,duan2023shifting}.
 
To leverage their respective strengths, the current advanced methods tend to combine different dimensions during confidence estimation.
Recent works~\cite{kuhn2023semantic,duan2023shifting} achieve outstanding performance on uncertainty estimation for open-domain question answering by combining logit-based approaches with semantics, using tools like bi-directional entailment or sentence encoders, aligning with Zone \textbf{2}. Rephrasing and round-trip translation can also be considered as using semantics to augment the remaining two methods~\cite{jiang2021can,zhao2023knowing}, corresponding to Zones \textbf{2} and \textbf{3}. P(True) leverages the self-evaluation capability of large language models~\cite{kadavath2022language}. While it primarily uses logit probability, it is clear that this probability is influenced by internal states and semantics, related to Zone \textbf{4}.
Anticipated advancements in collaborative information utilization will heighten computational demands, especially for nuanced semantic analysis~\cite{duan2023shifting}. This underscores the need for a careful balance between performance and resource efficiency.

    

\subsubsection{Black-box Methods} 
Black-box methods assume that all parameters during inference are unknown, allowing access only to the generations.
\paragraph{Linguistic confidence (verbalized method)} refers to prompting language models to express uncertainty in human language. This involves discerning different levels of uncertainty from the model's responses, such as "I don't know," "most probably," or "Obviously"~\cite{mielke2022reducing} or prompting the model to output various verbalized words (e.g., "lowest", "low", "medium", "high", "highest") or numbers (e.g., "$85\%$"). \citet{xiong2023can} demonstrated that prompting strategies like CoT~\cite{wei2022chain}, top-k~\cite{tian-etal-2023-just}, and their proposed multi-step method can improve the calibration of linguistic confidence.

\paragraph{Consistency-based estimation} assumes that a model's lack of confidence correlates with various responses, often leading to hallucinatory outputs. SelfCheckGPT~\cite{manakul2023selfcheckgpt} proposed a simple sampling-based approach that uses consistency among generations to find potential hallucinations. 
Five variants are utilized to measure the consistency: BERTScore~\cite{DBLP:conf/iclr/ZhangKWWA20}, question-answering, n-gram,
natural language inference (NLI) model~\cite{he2023debertav}, and LLM prompting. \citet{lin2023generating} proposed to calculate the similarity matrix between generations and then estimate the uncertainty based on the analysis of the similarity matrix, such as the sum of the eigenvalues of the graph Laplacian, the degree matrix, and the eccentricity. 

\paragraph{Surrogate models} \citet{shrivastava2023llamas} introduced white-box models as surrogate models, like LLaMA-2~\cite{touvron2023llama} and then employed logit-based methods to estimate the confidence of the target model when prompted with the same task. They also showed that integrating such confidence with linguistic confidence from black-box LLMs can provide better confidence estimates across various tasks.


\paragraph{Summary}
Figure~\ref{fig:black} illustrates the information sources for confidence evaluation when model states are not accessible: linguistic confidence, consistency, including lexical and semantic similarity, and surrogate models. Linguistic confidence can be elicited through prompts, but in practice, a mismatch between these has been observed~\cite{lin2022teaching,liu2023trustworthy}. Surrogate models~\cite{shrivastava2023llamas} facilitate white-box methods on black-box LLMs. However, they rely on the assumption of approximate parameter distribution of models, necessitating further work to validate their effectiveness. Consistency methods are computationally intensive but have proven effective in various tasks. They can benefit the remaining two approaches (Zone  \textbf{1} and \textbf{2}), such as the hybrid method proposed by \citet{xiong2023can}. Additionally, integrating all three methods (Zone \textbf{4}) has been explored by ~\citet{shrivastava2023llamas} to offer further benefits. Table~\ref{tab:estimation} presents the latest representative works in confidence estimation for large language models, briefly describing their proposed methods.


\subsection{Calibration Methods}
This section categories related work in terms of calibration objectives: to enhance the quality of generated text through calibration techniques and to improve the model's handling of unknown or ambiguous issues by enabling it to express uncertainty more accurately. The first half of Table~\ref{tab:genlms} presents recent work on calibrating LLMs over generation tasks.

\subsubsection{Improve the quality of generation}
Many studies~\cite{kumar2019calibration,wang2020inference,lu2022learning} indicated that the miscalibration of token-level logit probabilities during generation is one of the reasons for the decline in generation quality. \citet{kumar2019calibration} introduced a modified temperature scaling approach where the temperature value adjusts according to various factors, including the entropy of attention, token logit, token identity, and input coverage. \citet{wang2020inference} noted a pronounced prevalence of over-estimated tokens compared to under-estimated ones. They introduced \emph{graduated label smoothing}, applying heightened smoothing penalties to confident predictions. \citet{xiao2021hallucination} and \citet{van2022mutual} calibrated the token probability separately by adding a weighted uncertainty estimated with model ensembles~\cite{DBLP:conf/nips/Lakshminarayanan17} and pointwise mutual information between the source and the target tokens. \citet{zablotskaia2023uncertainty} adapted diverse methods to improve model calibration in neural summarization tasks.

\citet{zhao2022calibrating} suggested that MLE training can result in poorly calibrated sentence-level confidence, as the model is only exposed to one gold reference. They proposed the \textit{sequence likelihood calibration} (SLiC) technique to rectify this. It first generates $m$ multiple sequences $\{\hat{\mathbf{y}}\}_m$ from the initial model $\theta_0$, then calibrates the model's confidence with:
\begin{equation}
      \sum_{\{\mathbf{x}, \Bar{\mathbf{y}}\}} \mathcal{L}^{cal}(\theta, \mathbf{x}, \Bar{\mathbf{y}}, \{\hat{\mathbf{y}}\}_m) + \lambda \mathcal{L}^{reg}(\theta, \theta_0, \mathbf{x}, \Bar{\mathbf{y}})
\end{equation}
where the calibration loss $\mathcal{L}^{cal}$ aims to align models' decoded candidates' sequence likelihood according to their similarity to the reference $\Bar{\mathbf{y}}$, and the regularization loss $\mathcal{L}^{reg}$ prevents models from deviating strongly. They further introduced SLiC-HF~\cite{zhao2023slic}, which was designed to learn from human preferences.

\subsubsection{Improve the linguistic confidence}

\citet{mielke2022reducing} proposed a calibrator-controlled method for chatbots, which involves a trained calibrator to return the model confidence score and fine-tuned generative models to enable control over linguistic confidence. \citet{lin2022teaching} fine-tuned GPT-3 with the human-labeled dataset containing verbalized words and numbers to express uncertainty naturally. \citet{zhou2023navigating} empirically found that injecting expressions of uncertainty into prompts significantly increases the accuracy of GPT-3's answers and the calibration scores.  

Different datasets~\cite{amayuelas2023knowledge,yin2023large,wang2023not,liu2023prudent} have been presented on questions that language models cannot answer or for which there is no clear answer.~\citet{amayuelas2023knowledge} analyzed how different language models, including both smaller and open-source models, classify a dataset of various unanswerable questions. They observed that LLMs show varying accuracy levels depending on the question type, while smaller and open-source models tend to perform almost randomly. ~\citet{liu2023prudent} evaluated both open-source models like LLaMA-2~\cite{touvron2023llama}, Vicuna~\cite{vicuna2023}, and closed-source models such as GPT-3.5 and GPT-4, focusing on their refusal rate, accuracy, and uncertainty in handling unanswerable questions.

\section{LLMs for Classification Tasks}
\label{sec:gm4cl}

LLMs are recognized for their efficiency in classification tasks, enabling rapid task implementation via prompts~\cite{brown2020language,zhao2021calibrate}. Although the underlying principles of confidence estimation are similar to those in generation tasks, the objectives of calibration and the approaches differ significantly. 




\subsection{In-Context Learning}
In-context learning (ICL) is a new learning paradigm with LLMs, where the model learns to perform a task based on a few examples and the context in which the task is presented. Assuming that $k$ selected input-label pairs $(\mathbf{x}_1, y_1), \cdots, (\mathbf{x}_k, y_k)$ are given as demonstrations, with the predictive probability as the confidence, ICL makes predictions as follows:
    \begin{equation}
        \hat{y} = \argmax_{y} P(y|\mathbf{x}_1, y_1, \cdots, \mathbf{x}_k, y_k, \mathbf{x})
    \end{equation}
     When there are no demonstrations, the model performs zero-shot classification. 
     


\paragraph{Calibration methods}  We refer to the input-label pairs as $\mathbf{C}$ for context, and the original predictive probability is denoted as $P(y|\mathbf{C}, \mathbf{x})$. \citet{zhao2021calibrate} introduced a method called \emph{contextual calibration}. It gauges the model's bias with context-free prompts such as "$\texttt{[N/A]}$", "$\texttt{[MASK]}$" and an empty string. Then the context-free score is obtained by $\hat{\mathbf{P}}_{\texttt{cf}} = P(y|\mathbf{C}, \texttt{[N/A]})$. Subsequently, it transforms the scores with $\mathbf{W}=diag(\hat{\mathbf{p}}_{\texttt{cf}})^{-1}$ to offset the miscalibration.  \citet{fei2023mitigating} proposed \emph{domain-context calibration}, which estimates the prior bias for each class with $n$ times model average with random text of an average sentence length: $\Bar{\mathbf{P}}_{rd}(y|\mathbf{C}) = \frac{1}{n}\sum_{i=1}^n P(y|\mathbf{C}, \texttt{[RANDOM TEXT]})$. The prediction is obtained with: 
\begin{equation}
    \hat{y} = \argmax_{y} \frac{P(y|\mathbf{C}, \mathbf{x})}{\Bar{\mathbf{P}}_{rd}(y|\mathbf{C})}
\end{equation}

Some methods aim to improve few-shot learning performance by combining classic statistical machine learning techniques. \citet{nie2022improving} enhanced predictions by integrating a $k$-nearest-neighbor classifier with a datastore containing cached few-shot instance representations, while \citet{han2022prototypical} introduced \emph{prototypical calibration}, which employs Gaussian mixture models (GMM) to learn decision boundaries.

\subsection{ICL Application: Multiple-Choice Question Answering}
Multiple-choice question answering (MCQA) is an application of ICL, which is used in evaluating LLMs by prompting them to answer questions with predefined choices. The context $\mathbf{C}$ contains the question $\mathbf{q}$, and the set of options $\mathcal{I}(\mathbf{q}) = \{\mathbf{o}_1, \cdots, \mathbf{o}_K \}$, where each is prefaced with an identifier such as "A", and, if available, with a demonstration as an instruction. 

It is worth noting that implementing the evaluation protocols can significantly impact the ranking of models. For instance, the original evaluation of the MMLU~\cite{hendrycks2020measuring} ranks the probabilities of the four option identifiers. The answer is considered correct when the highest probability corresponds to the correct option. The HELM implementation~\cite{liang2022holistic} considers probabilities over the complete vocabulary. The HARNESS implementation\footnote{\href{https://github.com/EleutherAI/lm-evaluation-harness/tree/v0.3.0}{https://github.com/EleutherAI/lm-evaluation-harness/tree/v0.3.0}} prefers length-normalized probabilities of the entire answer sequence.

\paragraph{Calibration methods} \citet{jiang2021can} proposed various fine-tuning loss functions and temperature scaling for calibrating the performance of MQCA datasets. Additionally, they proposed techniques such as candidate output paraphrasing and input augmentation to calibrate the confidence. \citet{holtzman2021surface} claimed that surface form competition occurs when different valid surface forms compete for probability. Thus, they introduced \emph{domain conditional pointwise mutual information} (PMIDC), which reweighs each option according to a term that is proportional to its prior likelihood within the context of the specific zero-shot task. To overcome the bias from the choice position, \citet{DBLP:journals/corr/abs-2309-03882} proposed \textit{PriDe}, which first decomposes the observed model prediction distribution into an intrinsic prediction over option contents and a prior distribution over option identifiers and then estimates the prior by permuting option contents on a small number of test samples. \citet{kumar2022answer} believed that under the neutral context $\mathbf{C}_{\phi}$, the probabilities of different options should be the same, but obviously, the LLM cannot meet this condition, so they proposed using
$\log P(\mathbf{o}_k|\mathbf{C}) - sim(\mathbf{C}, \mathbf{C}_{\phi}) \log P(\mathbf{o}_k|\mathbf{C}_{\phi})$ to make the prediction. Given that $\mathbf{C}$ is very similar to the neutral context $\mathbf{C}_{\phi}$, the approach will assign an equal score to each choice.

\paragraph{Summary}The second half of Table~\ref{tab:genlms} lists recent calibration studies over classification tasks. Current calibration methods primarily aim to mitigate biases associated with labels or choice positions in MCQA~\cite{zhao2021calibrate,jiang2021can}. A growing trend in the field is to deepen the understanding of the ICL~\cite{holtzman2021surface} and to integrate semantics~\cite{kumar2022answer}. Besides, a systematic benchmark for evaluating different calibration methods is still missing.

\section{Applications }
\label{sec:applications}
Confidence estimation and calibration can be effectively employed in the following applications as an indispensable component in ensuring reliable AI.

\paragraph{Hallucination detection and mitigation}
Confidence or uncertainty can be applied as a signal for detecting and mitigating hallucinations generated by LLMs~\cite{zhang2023siren,huang2023survey}.
SelfCheckGPT \cite{manakul-etal-2023-cued} and $SAC^3$ \cite{Zhang2023SAC3RH} both explored hallucinations in the generation with self-consistency, while the latter also checked cross-model response consistency by taking generations from other models as the reference.
\citet{varshney2023stitch} proposed a method that leverages the model's logits to identify potential hallucinations, checks their correctness through a validation procedure, appends the repaired sentence to the prompt, and continues to generate.

\paragraph{Ambiguity detection and selective generation}
When identifying ambiguity in data or unanswerable questions, reliable LLMs are anticipated to refrain from providing answers rather than generating responses arbitrarily \cite{kamath2020selective}.
\citet{ren2022out} proposed a selective generation method based on relative Mahalanobis distance. \citet{zablotskaia2023uncertainty} provided a comprehensive benchmark study that evaluates various calibration methods in neural summarization.
\citet{cole2023selectively} and \citet{hou2023decomposing} respectively employed a disambiguate-and-answer approach and input clarification ensembling to measure data uncertainty for detecting ambiguous questions.

\paragraph{Uncertainty-guided data exploitation}
Through measuring data uncertainty, the most representative instances will be selected for few-shot learning \cite{yu2022cold} or human annotation \cite{su2022selective}.
Regarding the knowledge enhancement to LLMs, \citet{jiang2023active} proposed an adaptive multi-retrieval method that first forecasts future content and retrieves relevant documents stimulated by low-confidence tokens within upcoming sentences.

\section{Future Directions}
\label{sec:future_directions}

\paragraph{Comprehensive Benchmarks}
While confidence estimation and calibration have wide-ranging applications, a comprehensive benchmark across tasks and domains is required to better understand and evaluate these techniques' robustness and utility. Addressing this issue requires extensive human efforts to annotate the responses of LLMs, especially in long-form generation ~\cite{huang2024calibrating,mishra2024fine}.
Treating LLMs' long generation for confidence estimation and calibration by parts, instead of as a whole, offers a promising direction for further enhancement.

\paragraph{Multi-modal LLMs}
By employing additional pre-training with image-text pairings or by fine-tuning on specialized visual-instruction datasets, LLMs can be transited into the multimodal domain~\cite{DBLP:journals/corr/abs-2305-06500,DBLP:journals/corr/abs-2304-08485,DBLP:journals/corr/abs-2304-10592}. However, it remains unclear whether these confidence estimation methods are effective for multimodal large language models (MLLMs) and whether these models are well-calibrated. We look forward to more efforts in detecting hallucinations in MLLMs through confidence estimation and in calibrating these models to discern events that are impossible in the real world.


\paragraph{Calibration to human variation}
\citet{plank2022problem} clarified the prevalent existence of human variation, i.e., humans have different opinions when labeling the same data.
Human disagreement \cite{jiang2022investigating} can be attributed to task ambiguity \cite{tamkin2022task}, annotator's subjectivity \cite{sap2021annotators}, and input ambiguity \cite{DBLP:conf/acl/MeissnerTSA20}.
Recent work \cite{baan-etal-2022-stop, lee2023can} demonstrated the misalignment between LLM calibration measures and human disagreement in various learning paradigms.
Expressing the concern regarding different types of ambiguity \cite{xiong2023can}, abstaining from answering ambiguous questions \cite{yoshikawa2023selective}, and further resolving ambiguity \cite{DBLP:conf/acl/VarshneyB23} are necessary for trustworthy and reliable LLMs aligned with human variation.

\section{Conclusion}
This survey highlights the critical role of confidence estimation and calibration in addressing errors and biases in LLMs. The evolution of LLMs has paved the way for novel research opportunities and presented distinctive challenges. We first introduced the fundamental concepts of confidence and uncertainty, along with common metrics, estimation methods, and calibration techniques used in traditional discriminative models. We then identified the challenges these methods face in LLMs. Next, we delved into the latest research, introducing the principles, advantages, and drawbacks of various methods in generation and classification tasks. We concluded by discussing the current applications and future research directions. 

\section*{Limitations}

This survey mainly has the following limitations:

\paragraph{No experimental benchmarks} Without original experiments, this paper cannot offer empirical validation of the theories or concepts. This limits the paper's ability to contribute new, verified knowledge to the field.


\paragraph{Potential omissions} We have made our best effort to compile the latest advancements. Due to the rapid development in this field, there is still a possibility that some important work may have been overlooked.

\section*{Ethical Considerations and Potential Risks}
We anticipate no significant ethical concerns in our work. Our review surveys the latest developments in this research field, and as we did not conduct experiments, nor did we engage with risky datasets; we also did not employ any workers for manual annotation.




\bibliography{anthology,new}
\clearpage
\appendix
\section{Appendix}
\label{sec:appendix}

\begin{table*}[!htp]
\centering
\resizebox{\textwidth}{!}{
\begin{tabular}{p{4.5cm}p{5cm}p{6cm}p{6cm}}
\toprule
\textbf{Study} & \textbf{Model} & \textbf{Task} & \textbf{Calibration Methods} \\ 
\midrule
\cite{desai2020calibration}    &   BERT~\cite{devlin2018bert}, RoBERTa~\cite{liu2019roberta}    & \makecell[l]{Nature Language Inference, \\ Paraphrase Detection, \\Commonsense Reasoning }   & TS, LS   \\     
\cite{kim2023bag}     &   RoBERTa~\cite{liu2019roberta}  &  Text Classification   & \makecell[l]{BL, ERL, MixUp, DeepEnsemble,\\ MCDropout, MIMO   }                           \\ 
\cite{park2022calibration}      &   BERT~\cite{devlin2018bert}, RoBERTa~\cite{liu2019roberta}   & \makecell[l]{Nature Language Inference,\\ Paraphrase Detection, \\ Commonsense Reasoning}      &      TS, LS, MixUp, Manifold-MixUp, AUM-guided MixUp                    \\ 
\cite{zhang2021knowing}      &  \makecell[l]{BERT-based Span Extractor\\~\cite{zhang2021knowing}}     &  Extractive Question Answering    &       FBC                       \\ 
\cite{si2022re}      & \makecell[l]{BERT-based Span Extractor\\ ~\cite{si2022re} }    &  Extractive Question Answering   &      LS, TS, FBC     \\
\midrule
\bottomrule

\end{tabular}}
\caption{\textbf{Studies of discriminative LM calibration}. \textbf{Calibration methods:} LS=label smoothing, TS=temperature scaling, BL=brier loss, ERL=entropy regularization loss, BE=Bayesian Ensemble, SNGP: spectral-normalized Gaussian process, FBC=feature-based calibrator }
\label{tab:genlms1}
\end{table*}

\subsection{Confidence Estimation Methods}
The methods for confidence estimation have been extensively studied and can generally be categorized into the following groups:

\paragraph{Logit-based estimation }
Given the model input $\mathbf{x}$, the logit $\mathbf{z}$, along with the prediction $\hat{y}$ (i.e., the class with the highest probability emitted by softmax activation $\sigma$), the model confidence is estimated directly using the probability value:
\begin{equation}
    \texttt{conf}_{sp}(\mathbf{x},\hat{y}) = P(\hat{y}|\mathbf{x}) = \sigma(\mathbf{z})_{\hat{y}}
\end{equation}
There are methods for estimating confidence based on transformations of the logit probabilities, such as examining the gap between the two highest probabilities~\cite{yoshikawa2023selective} or utilizing entropy, which indicates the uncertainty with a larger value.


\paragraph{Ensemble-based \& Bayesian methods}

\emph{DeepEnsemble methods} \cite{DBLP:conf/nips/Lakshminarayanan17} train multiple neural networks independently and estimate the uncertainty by computing the variance of the outputs from these models. \emph{Monte Carlo dropout} (MCDropout, \citealt{DBLP:conf/icml/GalG16}) methods extend the dropout techniques to estimating uncertainty. 
As in the training phase, dropout is also applied during inference, and multiple forward passes are performed to obtain predictions.
The final prediction is obtained through averaging predictions, with the variability of the predictions reflecting the model uncertainty.

Methods such as deep-ensemble and MCDropout introduce a heavy computational overhead, especially when applied to LLMs \cite{malinin2020uncertainty, shelmanov-etal-2021-certain, DBLP:conf/acl/VazhentsevKSTTF22}, and there is the need to optimize the computation.
For example, determinantal point process \cite{DBLP:journals/ftml/KuleszaT12} can be applied to facilitate MCDropout by sampling diverse neurons in the dropout layer \cite{shelmanov-etal-2021-certain}.

\paragraph{Density-based estimation}  
Density-based approaches~\cite{lee2018simple,yoo2022detection} are based on the assumption that regions of the input space where training data is dense are regions where the model is likely to be more confident in its predictions. Conversely, regions with sparse training data are areas of higher uncertainty.
\citet{lee2018simple} first proposed a Mahalanobis distance-based confidence score, which calculates the distance between one test point and a Gaussian distribution fitting test data. The confidence estimation is obtained by exponentiating the negative value of the distance.

\paragraph{Confidence learning} employs a specific network branch to learn the confidence of model predictions.   \citet{devries2018learning} leveraged a confidence estimation branch to forecast scalar confidence, and the original probability is modified by interpolating the ground truth according to the confidence to provide ``hints'' during the training process. Additionally, it discourages the network from always asking for hints by applying a small penalty. \citet{corbiere2019addressing} empirically demonstrated that confidence based on true class probability (TCP) is better for distinguishing between correct and incorrect predictions. Given the ground truth $y$, TCP can be represented as $P(y|\textbf{x})$. However, $y$ is not available when estimating the confidence of the predictions. Hence, ~\citet{corbiere2019addressing} used a confidence learning network to learn TCP confidence during training.

\subsection{Model Calibration}
Calibration methods can be categorized based on their execution time as \emph{in-training} and \emph{post-hoc} methods.

\subsubsection{In-Training Calibration}
Research indicates that model generalization methods can be used for calibration~\cite{kim2023bag}, and calibration methods can enhance model performance, particularly in out-of-domain generation~\cite{desai2020calibration}.
\paragraph{Novel loss functions} Many studies considered the \textit{cross-entropy} (CE) loss to be one of the causes leading to model miscalibration~\cite{mukhoti2020calibrating,kim2023bag}.  \citet{mukhoti2020calibrating} demonstrated that \emph{focal loss}~\cite{lin2017focal}, designed to give more importance to hard-to-classify examples and to down-weight the easy-to-classify examples, can improve the calibration of neural networks. The \emph{correctness ranking loss} (CRL; \citealt{moon2020confidence}) calibrated models by penalizing incorrect rankings within the same batch and by using the difference in proportions as the margin to differentiate sample confidence. Besides, \emph{entropy regularization loss} (ERL; \citealt{pereyra2017regularizing}) and \emph{label smoothing} (LS; \citealt{szegedy2016rethinking}) were introduced to discourage overly confident output distributions.

\paragraph{Data augmentation} involves creating new training examples by applying various transformations or perturbations to the original data. It has been widely used for calibration of discriminative LMs by alleviating the issue of over-confidence, such as MixUp~\cite{zhang2017mixup}, EDA~\cite{wei2019eda}, Manifold-MixUp~\cite{verma2019manifold}, MIMO~\cite{mimo2021} and AUM-guided MixUp~\cite{park2022calibration}.

\paragraph{Ensemble and Bayesian methods} were initially introduced to quantify model uncertainty. However, both can also be valuable for model calibration, as they can enhance accuracy, mitigate overfitting, and reduce overconfidence~\cite{kong2020calibrated,kim2023bag}.

\subsubsection{Post-Hoc Calibration}
\paragraph{Scaling methods} are exemplified by \emph{matrix scaling}, \emph{vector scaling} and \emph{temperature scaling}~\cite{Guo2017OnCO}.
Using a validation set, they fine-tune the predicted probabilities to better align with the true outcomes, leveraging the \emph{negative log-likelihood} (NLL) loss. 
Among them, temperature scaling (TS) is popular due to its low complexity and efficiency. It involves re-weighting the logits before the softmax function by a learned scalar $\tau$, known as the \emph{temperature}. 


\paragraph{Feature-based calibrator} leverages both input features and model predictions to refine the predicted probabilities. To train the calibrator, one first applies a trained model on a validation dataset. Subsequently, both the original input features and the model's predictions from this dataset are passed to a binary classifier~\cite{jagannatha2020calibrating,jiang2021can,si2022re}.

\subsection{Summary} 
\paragraph{Confidence estimation} Logit-based methods stand out as the most straightforward to implement and interpret. Reducing computational cost and improving the sampling efficiency pose challenges to ensemble-based and Bayesian methods. Density-based estimation can be used to identify which data points are associated with different types of uncertainties. However, it requires assumptions of data distribution~\cite{baan2023uncertainty} and can also be computationally intensive when dealing with large datasets~\cite{sun2022daux}. Confidence learning can acquire task-relevant confidence; however, it requires modifying the neural network and performing specific training.

\paragraph{Model calibration} Post-hoc methods are generally model-independent and can calibrate probabilities without impacting the model's performance~\cite{Guo2017OnCO}. \citet{desai2020calibration} empirically found that temperature scaling effectively reduces calibration error in-domain, whereas label smoothing is more beneficial in out-of-domain settings. \citet{kim2023bag} found that augmentation can enhance both classification accuracy and calibration performance. However, ensemble methods may sometimes degrade model calibration if individual members produce similar predictions due to overfitting. Table~\ref{tab:genlms1} represents significant work in calibrating discriminative LMs. We have comprehensively listed the models, tasks, and calibration methods they employed.


\end{document}